\documentstyle[a4,11pt]{article}

\title{Natural language processing:\\
she needs something old and something new\\
(maybe something borrowed and something blue, too)}

\author{Karen Sparck Jones\\
\phantom{FILLER}\\
Computer Laboratory, University of Cambridge\\
New Museums Site, Pembroke Street, Cambridge CB2 3QG, England\\
{\em ksj@cl.cam.ac.uk}}

\date{}

\begin{document}

\maketitle

\begin{center}
{\em Presidential Address, June 1994, Association for Computational
Linguistics}
\end{center}

\begin{abstract}
   Given the present state of work in natural language processing, this
address argues first, that advance in both science and applications
requires a
revival of concern with what language is about, broadly speaking the world;
and second, that an attack on the summarising task, which is made ever more
important by the growth of electronic text resources and requires an
understanding of
the role of large-scale discourse structure in marking important text
content,
is a good way forward.
\end{abstract}

   I want to assess where we are now, in computational linguistics and
natural language processing, compared with where we started, and to put my
view of what we need to do next.
We should not cut off the past too soon, but keep a sense of perspective
so that we can properly judge what advance we are making, i.e. whether
progress is real or illusory, forward or sideways and, if real, how good
relative to our desired end point and not just our beginning one.

   Computational linguistics, or natural language processing
(NLP), is nearly as old as serious computing.
Work began more than forty years ago, and one can see it going through
successive phases, roughly ten year periods from the late fifties onwards.
I have discussed these phases more fully elsewhere (KSJ94), so I will say
only
enough about them here to provide a context for my later argument.

   The first phase, beginning in the late fifties, was linguistically
oriented,
focusing on machine translation, with people learning, painfully,
how to do things computationally.
The second phase, from the late sixties to the late seventies, recognised
the
role of real world knowledge, was strongly motivated by AI, and drove NLP
from
this.
The third phase, dominating the eighties, acknowledged the specific
modulating
or controlling function for language relative to the world, and tried to
capture this, in its necessarily systematic aspect, in grammatico-logical
models for NLP.
The fourth phase, that we are in now, while taking the grammatico-logical
skeleton for granted, recognises the significance of actual
language usage, both idiosyncrastic and habitual, as a constraint
on performance, and is therefore heavily into data mining from corpora.

   All of these phases have contributed something to the growth of knowhow.
But what can we actually do now, given NLP's necessary concerns both
with generic capabilities like syntactic parsing and with particular
tasks like translation, i.e. with both subsystem and whole system functions?

   What I see as most significant are the following.

   First, we have engines, reasonably solid generic systems with decent
coverage,
that we can put to work in interpretation to deliver semantic
representations
for sentences or, in generation, to deliver sentences from representations:
for example SRI, BBN, NYU or ISI's systems.
We have, that is, more than subfunction components: we have respectably
engineered multi-component systems.

   Second, we can build tightly-targeted application systems for specific
purposes, for example for finding index terms in open text,
or for extracting data from banking telexes.

   Third, we can do this not only because we have engines to direct
towards tasks and supporting tools to deploy, e.g. for lexical
acquisition, but because we now have some understanding, based
on experience, of how to design systems suited to particular tasks:
for instance we can judge what level of syntax analysis is appropriate.

   Fourth, moreover, and as a very broad generalisation, we often find that
what we need to work with as meaning representations for sentences are some
sort of predicate-argument structures with case roles, that will connect
with world knowledge similarly represented, though the details both
of the procedures through which we arrive at these representations and
of the form of the representations themselves (e.g. frames, networks, etc)
can vary without ill effects.
We can thus, when we are designing systems for individual tasks, see them as
running our engines under different sets of constraints, or with more or
less tolerance.

   Fifth, and finally, we have recently begun to reach out into spoken
language systems where speech processing and language processing are
at least to some extent genuinely interlinked, and are not just done by
two separate boxes, butted together.
Getting into spoken language processing, not just speech processing,
is both important in itself and is evidence of having some solid NLP
ground to stand on.

   All this seems like cause for satisfaction: there is no doubt that
we have made some progress not only when viewed from our starting point
but from our finishing one - i.e. having systems with something like
human capabilities, - and that we can do some of the things we wanted
and needed to do when work in the field began.

   However when we look more carefully things are not so encouraging.

   Even if we think we're in the business of pure science, i.e. of
computational
modelling of language processing for its own sake, there's so much we
haven't
done, for instance in discourse processing.
More importantly, whether we take practical NLP system building as the
real validation for modelling or want systems for themselves, we can see,
if we take the market place as an indicator, that we've not got
too far.
Compared with the twin playing with us in the toddlers' pen forty years
ago, namely computing, we're nowhere.
There are computers everywhere - in offices, factories, shops, hospitals,
homes and boats, - but in spite of the fact that natural language
is {\em our} medium we don't have NLP everywhere, indeed we hardly have it
anywhere to speak of:
lots of Dragon Dictates have been sold, but they don't do NLP; lots of
spellcheckers are sold, but they don't do NLP; lots of Q\&As were sold,
but who uses the NLP bit, and it's fairly modest anyway.

   When we look for NLP systems worthy of the name out there doing a job,
there
are not many, and those there are, like Systran as used by Rank Xerox at
Welwyn,
are very thoroughly tailored to specific applications.
Some Systran implementations at least show the importance of gritty language
detail, but these and other operational systems, e.g. for message handling,
tend to have weak or partial models of language processing and very limited
models of their tasks.
In fact these systems are typically working with models of the particular
versions of their tasks associated with individual applications.

   Why haven't we done better?
Is it that the enterprise is intrinsically tough, or that we've gone wrong
somehow?

   I think we can get insights if we look at what we found when we attempted
some tasks, and can then get pointers for where to go next.

   I shall take database query as the first case because everyone,
from an early stage in NLP research and notably in the seventies,
thought it both useful and a doddle.
In fact it wasn't a doddle, for reasons now fairly familiar.
This is because you have to build a non-trivial domain model both
to link language and data and to cope with language input that's astray,
and even so it's hard to keep the user within bounds, i.e.
there is usually such a mismatch between front end NLP requirements and
back end lack of power that you finish up building a complete inquiry
system as an application-tailored whole, with a domain model
at its core that has to be both more comprehensive than the basic
data model and capable of supporting non-trivial inference.
For example, given a town planning database about lots
with attributes like owners and values, we need such a domain model both to
handle proper
but indirect questions like ``Who owns Market Street?" (i.e. Who are the
owners
of the lots on Market Street), and to respond cooperatively to questions
that are outside the scope of the database, like (say) ``Who is
developing Market Street?"
We can to some extent manage the database case by now, but it is an effort
and requires more application tailoring than originally envisaged.

   Or, taking translation as the second example, while one can win with
Meteo and, in some cases, not do badly with a purely linguistic
approach, it is all too easy to come unstuck through lack of world
knowledge.
There are well-known problems, for instance, in the area of
intersentential anaphoric references, as with ``This" in
``It's hard to know what to do about the exam schedule. This is
something we've really got to overcome", which refers to not knowing, and
not
to the exam schedule.
However with unrestricted subject-matter text one is in real difficulty
about providing a world model to underpin processing.

   Or, taking the presently fashionable message-processing task as a
third illustration, we can currently only do this if we know, quite
specifically, in advance, what sorts of data items we are looking
for, e.g. facts about earthquakes, and even then may need further
support from domain modelling,
for example to relate events in successive messages about a
traffic accident.

   When we contrast these cases with others like document indexing and
retrieval,
which we can do quite well, we see very clearly that where the task
requires only shallow NLP, as in retrieval where statistically-based
content indexing is very shallow, we can have a successful general
approach that carries across individual applications without any requirement
for crafting.
Whereas when the task requires deeper NLP, while we may have some
component generality, e.g. in syntax, we need a great deal of
crafting.
This does not mean, either, that there are no real supporting models in
the wide, shallow cases: there are perfectly good and effective
language and task models.
The lack of models is for the deep cases.

   All of this is familiar enough: is there, then, a rational research
strategy that will promote advances for the more difficult and central
tasks requiring deep language processing, where we need not merely
better individual systems than we have now, but more generalisation
across task instances?

   The currently favoured route forward is, as I noted earlier, via
information derived from corpus analysis: for example we can use
frequency or collocation data as a basis for preferring one word
sense to another.
It is clear that usage data is valuable,
\footnote{quite apart from my liking for the revival of
interest in my own first research topic, semantic classification}
and that the linguistic patterns identified
in corpora encapsulate facts about things and their relations in the
world.
But there is still much to do in this area.
We need better corpora, not only in the obvious sense of ones big
enough to give us information about rare phenomena, but also in the
less obvious sense of corpora representative enough to give us reliable
information about any phenomena, for instance about genres.
We also need better methods of data analysis, i.e. classification methods,
especially for semantic information, that are richer and more
appropriate than just similarity computation or hierarchic
clustering, specifically because they allow classes to overlap.
We need, that is, in relation to present corpus interests, both
to design our corpora more carefully, and to investigate our analysis
methods more deeply, than hitherto, as well as to evaluate the results
very thoroughly, through their use for NLP.

   However even if we proceed with this, it does not address the problems
I believe we really have, and should face up to.
So while it is important that we should continue to consolidate
on what we can already do in system building, pushing outward everywhere
to get more power and coverage, and that we should also develop and
exploit corpus-based linguistic resources, we need to complement
these activities by addressing other concerns.
One of these concerns
is a core issue for language processing in general that we are
currently, and damagingly, neglecting; and the other is a family
of tasks that are both important in their own right and either
depend on an attack on the core issue or offer a valuable study context
for this.

   So what is the core problem?

   It is a perfectly familiar one, namely how language and the world are
related.
In one sense it is an issue with which we are engaged whenever we try
to do NLP.
But at the same time we are not currently, I believe, facing up to
it squarely.
We must remind ourselves, all the time, that people use language to say
something about something: they have messages, referring to things
(in the broadest sense), that they want to convey.

   Now when I say we are not facing up to this, I mean the following.
Suppose I say:

``The cat looks pretty contented, sitting there on the mat".

\noindent
For any of a range of possible responses, e.g.

``I wouldn't have said so"

\noindent
or

``So it ought, it's just eaten a large rat''

\noindent
a system has to know a lot about the world, to be able to
use that knowledge (inferentially), and to be able to communicate it
to intended effect.

   Of course there is nothing new in this; and we are all aware of
how hard it is to get systems able to do what is required.
My point is rather that we will not advance unless we put this
problem up front, and stop fudging it in the way we are currently
doing.
Relying on corpus-based linguistic data, or domain-specific semantic
patterns, is trying to finesse direct reference to the world, but
is restricting discourse in the process.
On the other hand, providing domain models but concentrating only on
limited applications, and supplying only the minimal models
deemed necessary to support primary language processing, is marginalising
world knowledge.
(In the current versions of the example tasks mentioned earlier we often
find, even when we do have some sort of domain model, such minimalist
approaches.)
Again, allowing dialogue participants to have beliefs, goals and plans,
but in specific task-oriented contexts, is making the relation between
language function and reference too tight and so too simplistic.
In other words, by arguing, after the last AI-motivated period in NLP
during the seventies, that
language knowledge should have a bigger role, we have given world
knowledge and purposes too small a role.

   This is not surprising, because embarking on CYC-like endeavours is
daunting and risky.
However unless we keep asking what people use language for, and
try to provide our systems with similar capabilities, their
growth will be stunted.
Is there, therefore, anything in our particular situation now, that can
give us a handle on this?

   I believe that there is something new that can stimulate work on our core
problem in a productive way, as follows.

   Everyone is becoming aware of the masses of language, and especially
text, material, flooding onto the networks, material that is actually
just stuff, and only potentially information.
Coping with this as a single, human end-user presents all kinds of
problems under the heading of information characterisation and retrieval,
for example how to achieve sufficiently discriminating text or
subtext selection.
However something more powerful is needed
than either whole text, or even subtext, selection.
What is needed is whole text
condensation, i.e. summarisation.
Summarisation identifies the key content of a whole text, but provides it
with context that a selected subtext can lack.

   Summarisation has always, as a key human capability, been a challenge for
NLP, but hitherto one that has hardly been attempted.
Now, I believe, we have really got to tackle automatic summarising, and
with much better kinds of approach than those tried so far.
We don't want just surface sentence extraction: this is general, but far too
crude.
But we don't, either, want present message understanding-type methods
(as exemplified by MUC), not just because these are domain and even
application specific, but because they are prescriptive on what will be
deemed important content for any individual text.
We rather need methods that have source texts supply their own
important content.
Texts are, in general, individual, and each has its own message to convey.
We thus want summarising systems that are responsive, not pre-emptive.

   It is also clear that for summarisation we have to treat source texts as
wholes,
since what is important is a function of the whole; and
in consequence we must look at the large-scale text structure that is
the means of organising text content for communicative purposes.
There is no doubt that there is large scale structure, but there are
many views about the nature of this discourse or text structure, some
at least well known to the NLP community, though others are less
familiar.
The point to note, however, is that while
different accounts may deal with distinct types of information,
these are all necessary to a discourse, as discourse, even though they may
have
quite different organisational structures.

   Thus we have linguistic information and a linguistic structure, for
instance
of sentence or paragraph parallelism;
we have domain information and its structure, for instance categorising
objects or linking events; and we have communicative information
and a communicative structure, for instance reflecting the aim of
convincing through
illustrative examples.
In relation to summarising, initial studies we have done in
Cambridge, for a small set of test texts, show both that there are
large-scale structures of these types and that they are distinct - often
very distinct - from one another
(though they are of course also related).
At the same time, they all supply indications of what is important in a
source text, and so of what we should seek to capture to constitute the body
of its summary (KSJ93).

   For my present purpose, therefore, the key point is that a summary is
primarily
concerned with what a text is essentially about.
So while the linguistic properties of a text may mark and help to
convey this, summarising also requires operations on text content,
i.e. ones deploying world knowledge and invoking inference and
further,
operations dealing with the specially important aspects of the world having
to do with communicators' intentions.

   I am not claiming that no-one has ever thought about the role of
large-scale
discourse structure, just as I am not maintaining that
no-one has ever thought about summarising.
Nor am I saying that we should address new problems before we have
solved old ones.
What I am saying is that summarising is both a critical NLP function
and an increasingly pressing NLP task, and that addressing it in any
general way will force us to make what texts are about, and
why they are about this, a central concern for NLP.
We need to draw the veil of language aside, and summarising will
compel us to do this.
Summarising will also, I believe, have benefits for NLP in forcing us
to consider extended discourse: it is too easy to get fixated on
individual sentences, or sentence pairs in a dialogue turn, and not see the
wood for
the trees.

   Thinking about summarising under the sort of classical definition I
have used is, however, only part of what the electronic future involves,
whether as need or opportunity.
When we reflect on all the implications of reaching information we
want or need in all the vast volumes of material that are becoming
available,
this suggests, even implies, we are going to need NLP-based capabilities
that we have not really needed to the same extent, and have barely
examined, so far.
We want the ability to change both the grain and type of information
representation, and to be able to move from one to another at will,
instantly.
That is, we want to be able to manage access to, and use of, material
in these large files by having a variety of abbreviated representations
of the fuller sources to work with, to meet different needs, where these
representations may either already exist or be constructed under present
context constraints.
We of course form and apply brief
representations ourselves, as individual information seekers and users.
What we have now is a system requirement for these construction
and exploitation processes to work on a scale
and with a speed, and potentially also in a dynamic and tailorable way,
that is something new.

   Summarising thus has a critical role, especially if we now think
more broadly of the many other forms and levels of reduced text
representation
there may be, for example:

a few index keys constitute a minimal indicative summary;

a title provides another tight condensation;

a selected passage reduces the source to a salient part; while

some extracted facts treat key information in a different, reorganising or
normalising,\\
\phantom{\ \ \ \ \ }way.

\noindent
We can equally, for any one of these, have fuller or sparser versions,
for example phrasal or single word keys, titles with or without modifiers
on nominals, more or less extensive or detailed extracts.

   We are familiar, in ordinary information processing or seeking,
with delivering or using a few of these
alternatives either concurrently or conjointly, e.g. titles and keywords.
I also talked, in 1983 (KSJ83), about shifting from one meaning
representation
to another, in that case between data and document retrieval derivatives
of the same natural language question.
I am thinking here of a much richer version of the same idea, where we
want to be able to supply different brief surrogates for extended
sources (or even sets of these), to meet different information-seeking
needs, for example:

for choosing a few sources, given many;

for skimming many sources;

for illustrating some sources;

for replacing several sources (as in a conventional database).

   In providing these surrogates we have of course to start from the
sources, but once we are underway we can use the fact that we are
dealing with language objects to establish organisational links
between surrogates themselves, and to derive new ones, even where the
languages involved are not identical.
In the enterprise as a whole we may sometimes be able to get most of
our processing done by primarily linguistic means, for example
by statistical techniques; but in other, more exigent cases, we
may have to rely heavily on world knowledge.
Thus in this extended NLP task area, we shall again have to engage with
what texts are about, and can therefore also look to take the benefits
back to other tasks we are already working on.
At the same time, we are seeking far more than mere hypertext links,
since we are forming explicit, if brief, discourses.

   So much for my vision of the future: a sketch of what I shall call
the information chameleon.
Heaven knows how to do it all, but we have still, I firmly believe, to try.
   However, in conclusion and looking in a more definitely down to earth
direction, I should note the concomitant evaluation challenge, and hence the
need for a proper evaluation methodology.
The current eveluation binge may not be to everyone's
taste, but I believe we have gained enormously from being pushed into
taking seriously something we should have addressed more thoroughly long
ago.
Summary evaluation for any of the forms mentioned is going to be really
hard.
But it will be important for us in NLP research because it will
compel us to take account both of system parameters and environment
variables - in this case what people use summaries for as well as
what their sources are like; - and it will therefore help us to avoid the
danger of improperly divorcing NLP systems from their contexts.

\vspace{15mm}

{\bf References}

   KSJ83: K. Sparck Jones,
``Shifting meaning representations",
{\em Proceedings of the Eighth International Joint Conference on
Artificial Intelligence,} 1983, 621-623.

   KSJ93: K. Sparck Jones,
``What might be in a summary?",
{\em Information Retrieval 93: Von der Modellierung zur Anwendung}
(Ed. Knorz, Krause and Womser-Hacker), Konstanz: Universitatsverlag
Konstanz), 1993, 9-26.
(//ftp.cl.cam.ac.uk/public/papers/ksj/
ksj-whats-in-a-summary.ps.gz)

   KSJ94: K. Sparck Jones,
``Natural language processing: a historical review",
in {\em Current Issues in Computational Linguistics: in Honour of Don
Walker}
(Ed. Zampolli, Calzolari and Palmer), Amsterdam: Kluwer, 1994.

\end{document}